\documentclass{INTERSPEECH2023}

\interspeechcameraready

\title{Unsupervised Auditory and Semantic Entrainment Models with Deep Neural Networks}
\name{Jay Kejriwal$^{1,2}$, Štefan Beňuš$^{1,3}$, Lina M. Rojas-Barahona$^{4}$}
\address{
  $^1$Institute of Informatics, {Slovak Academy of Sciences}, {Slovakia} \\
$^2$Faculty of Informatics and Information Technology, {Slovak Technical University}, {Slovakia}\\
  $^3$Constantine the Philosopher University, {Nitra, Slovakia}\\ 
  $^4$Orange Innovation, {Lannion, France}}
\email{jay.kejriwal@savba.sk, sbenus@ukf.sk, linamaria.rojasbarahona@orange.com}

\begin{document}

\maketitle
 
\begin{abstract}

Speakers tend to engage in adaptive behavior, known as entrainment, when they become similar to their interlocutor in various aspects of speaking. We present an unsupervised deep learning framework that derives meaningful representation from textual features for developing semantic entrainment. We investigate the model's performance by extracting features using different variations of the BERT model (DistilBERT and XLM-RoBERTa) and Google's universal sentence encoder (USE) embeddings on two human-human (HH) corpora (The Fisher Corpus English Part 1, Columbia games corpus) and one human-machine (HM) corpus (Voice Assistant Conversation Corpus (VACC)). In addition to semantic features we also trained DNN-based models utilizing two auditory embeddings (TRIpLet Loss network (TRILL) vectors, Low-level descriptors (LLD) features) and two units of analysis (Inter pausal unit and Turn). The results show that semantic entrainment can be assessed with our model, that models can distinguish between HH and HM interactions and that the two units of analysis for extracting acoustic features provide comparable findings. 
\end{abstract}

\section{Introduction} 
Entrainment in spoken interaction is the tendency of speakers to adjust some properties of their speech to match the characteristics of their interlocutors. It affects several linguistic dimensions, such as lexical choice \cite{brennan1996a}, syntactic structure \cite{reitter2006a}, acoustic prosodic features \cite{levitan2011a}, or semantic similarity \cite{ta-a}. In addition, it correlates with different social aspects of the conversation, such as task success \cite{reitter2007a}, liking \cite{ireland2011a}, cooperation \cite{manson2013a}, or naturalness and rapport \cite{lubold2015a}. 

Assessing and formally modelling entrainment in both human-human (HH) and human-machine (HM) spoken interactions is widely researched particularly due to the assumption that implementing natural entrainment functionalities reliably detected in HH dialogues would increase the naturalness and efficacy of applications using HM spoken interactions. To this end, several methods for measuring entrainment have been used ranging from time-series analyses, Pearson’s correlations, recurrence analyses, and spectral methods \cite{delaherche2012a}. Most of these approaches, however, assume a linear relationship between features of adjacent speaker turns. Yet, current understanding of entrainment is that it is a complex multi-layered phenomenon. For example, in \cite{p2016a}, the authors reported that also dis-entrainment could sometimes enhance conversation development. It seems that linear functions cannot capture this diverse and complex nature of entrainment. 

A potential approach addressing this issue was suggested in \cite{nasir2018a} who developed a DNN-based Neural entrainment distance (NED) that uses a non-linear function to measure auditory entrainment. They developed a novel unsupervised model, which relies on an auto-encoder architecture to produce the next turn's features. Instead of compressing and reconstructing the original embeddings, the main goal of this architecture was to learn the representation of the next turn based on the previous turn. Using these bottleneck features, NED represents the degree of auditory entrainment. With low-level description (LLD) auditory features extracted from inter-pausal units (IPUs) they reported classification accuracy of 98.87\% on Fisher corpus English Part 1 \cite{cieri2004a}. In their extended work, \cite{nasir-a} proposed a triplet network-based approach where they used \textit{i}-vectors for training the DNN model measuring auditory entrainment. They compared the performance of both approaches on different corpora and reported that the NED-based approach with LLD features provides the highest accuracy of 98.87\% in the Fisher corpus but the triplet network-based approach with \textit{i}-vectors trained on Fisher corpus was best (94.63\%) in the Suicide Risk Assessment corpus \cite{bryan_associations_2018}. Finally, \cite{weise_decoupling_2020}, proposed a DNN based entrainment model that isolates the effect of consistency; i.e. the tendency to adhere to one's own vocal style. The authors extracted LLD features from each adjacent IPU of both speakers and trained the DNN model using the deconfounding measures proposed in \cite{pryzant}. They reported that their model performs slightly worse than the NED measure, with an accuracy of 92.3\%.

In this paper we have two main goals in extending this promising line of work. While the mentioned studies explore the auditory modality of entrainment, to fully understand HH entrainment and design applications using it in HM interactions, also the textual modality has to be included. Hence, our first goal is to propose augmenting the original auto-encoder NED-based DNN architecture so that it could be used to assess also semantic entrainment. We test our approach and the performance of the models using variations of BERT and Google's Universal Sentence Encoder (USE).

The second goal is to test some aspects of the auditory NED-based approach in the effort to improve our understanding of its usability and reliability. First, it is not clear how different features, embeddings and units of analysis affect the performance of these models. Several types of auditory features have been identified for detecting auditory entrainment, including LLDs comprising temporal, spectral, and acoustic-prosodic features or DNN embeddings such as \textit{i}-vectors. Also, TRILL vectors have shown state-of-the-art performance in a classification task related to detecting stress in speech outperforming LLDs and also the above-mentioned \textit{i}-vectors \cite{kejriwal2022a}. Therefore, we test these types of features with the NED-based architecture on three different spoken corpora. Regarding the unit of analysis for extracting the features, IPUs are more common in this research but the entire turns might provide richer representations of speaker's characteristics. Hence, we designed experiments testing the performance of LLD and TRILL features on NED-based auditory entrainment models on three datasets using two different units of analysis.

Second, in the original papers describing the development of the NED-based model, cross-validation is done using the hold-out method, with data split into 80:10:10 for training, validation, and testing respectively. However, k-fold cross-validation, i.e. randomly splitting the dataset into ‘k’ groups, have been shown to provide more consistent results on smaller and larger datasets with quality classification than the hold-out \cite{yadav_analysis_2016}. Hence, it is warranted to evaluate the performance of the NED-based models using 10-fold cross-validation.

Finally, given the broader goal of developing entrainment functionalities in spoken HM interactions, it is critical to have robust and reliable measures of entrainment in both HH and HM scenarios of various domains and genres. We thus employ two HH corpora (The Fisher Corpus English Part 1, Columbia games corpus) and one HM corpus (Voice Assistant Conversation Corpus, VACC)) to test if the NED-based entrainment measure can safely distinguish between HH and HM interactions.

In sum, we propose adjustments to the auto-encoder architecture for using NED-based approach also for semantic entrainment, and we compare the performance of DNN-auditory entrainment models by training them on different features and units of analysis using three different corpora to understand if entrainment can be captured in HH and HM interactions. We also compare the performance of DNN-auditory and semantic entrainment models using different auditory and semantic embeddings  by splitting the dataset with 10-fold cross-validation, which reduces the variance. The experimental results show that the auditory NED model provides better accuracy with TRILL vectors, and XLM-RoBERTa embeddings provide better accuracy on the semantic entrainment model. This implies that non-semantic speech representation (DNN embeddings) for speech and textual embeddings can be useful for speech and semantic entrainment detection in a uni-model architecture.

\section{Data and features} \label{Data and features}

\subsection{Datasets}

\hspace{\parindent} \textbf{The Fisher Corpus English Part 1} \cite{cieri2004a} consists of 5850 spontaneous telephone dyadic conversations between native English speakers. During each session, two previously unacquainted subjects discuss a topic for 10 min. The dataset has 984 hours of speech, contains manual transcripts with time stamps for the speakers' turn and pauses enabling us to extract turns and IPUs. 

\textbf{Columbia Games Corpus} \cite{hirschberg2021a} contains 12 dyadic conversations between native speakers of Standard American English. Six females and seven males participated, 11 participated in two sessions on different days. The dyads played four computer games, requiring verbal collaboration. A curtain ensured only verbal communication. Over 9h of recordings were made. 

\textbf{Voice Assistant Conversation Corpus (VACC)} \cite{VACC} includes recordings of 27 native German speakers. An Amazon Echo Dot 2nd generation was used to explore the interactions between human-computer (solo condition) and human-human-computer (confederate condition). In the interaction, Calendar (formal) and Quiz (informal) tasks were used. In the Calendar task, the participant scheduled several appointments with the confederate during predefined weeks by interacting with Alexa. In the Quiz task, the participant answered trivia questions, such as “When was Albert Einstein born?” In total, approximately 13,500 utterances in over 17h was recorded (31 minutes average per interaction). Information about speakers and turn times was manually annotated. In the current study, we used the solo condition only.

\subsection{Feature extraction}

\hspace{\parindent} We extracted LLDs and TRILL vectors from the three datasets. For LLDs we used the OpenSMILE toolkit \cite{eyben2010a}. LLDs involve 38 features comprising 4 prosody ones (pitch, energy, their deltas), 31 spectral ones (15 MFCCs, 8 MFBs, 8 LSFs), and 3 voice quality ones (shimmer, two variants of jitter). Furthermore, we extracted 6 functionals of each acoustic feature (mean, median, standard deviation, 1st percentile, 99th percentile, and range (99th percentile - 1st percentile). We extracted 38 $\times$ 6 = 228 LLD features for each unit (IPU or turn). Step size and time frame were set at 25ms and 10ms, respectively. In addition, we \textit{z}-score normalized the features. We used the TRILL vector \cite{hoffer2018a} on all three datasets to extract each unit's auditory embeddings representing 512 one-dimensional vectors. 

Furthermore, we extracted semantic features using three different neural network-trained models. For the two English corpora we used the fine-tuned DistilBERT \cite{reimers2019a}, where each turn is encoded into 768 one-dimensional semantic features. For the German dataset (VACC), we utilized XLM-RoBERTa \cite{germanbert}, where each unit is encoded into 768 one-dimensional vectors. For comparison, we also extracted semantic features from all three datasets using Google's neural network-trained model Universal Sentence Encoder (USE) \cite{cer2018a}, representing 512 one-dimensional semantic features for each unit.

\subsection{Modelling with DNN}

\subsubsection{Auditory modelling}
\begin{figure}[h]
\centering
\includegraphics[width=80mm]{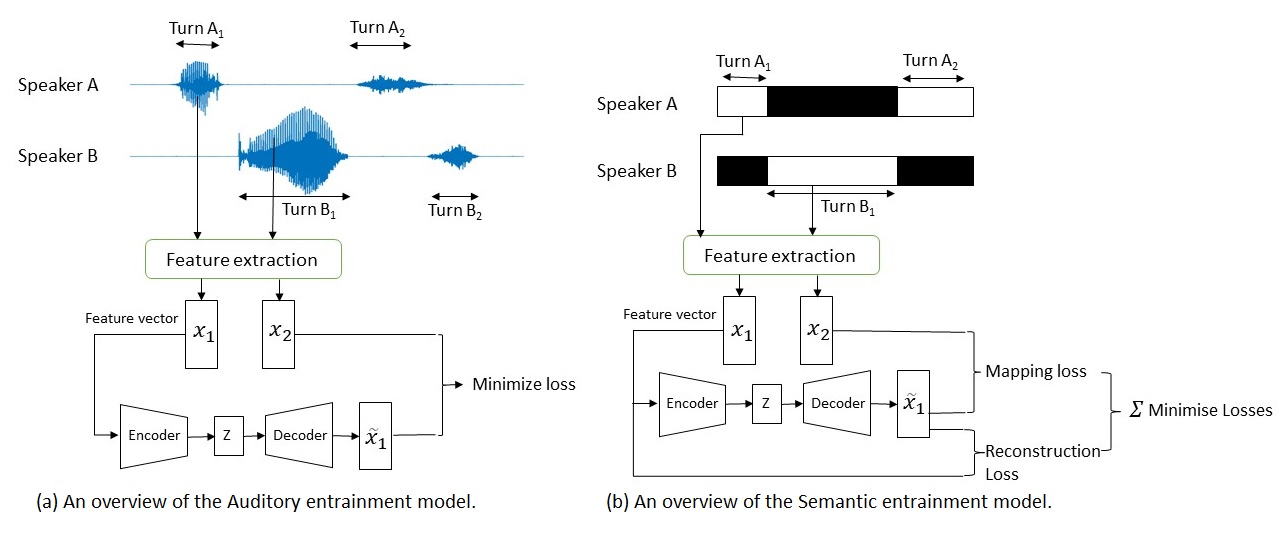}
\caption{An overview of (a) Auditory and (b) Semantic entrainment models}
\label{fig:DNN_aud_sem}
\end{figure}

\hspace{\parindent} We utilized the neural architecture as proposed in \cite{nasir2018a} illustrated in Figure \ref{fig:DNN_aud_sem} (a). As a first step, ${x_1}$ is the input to the encoder network. By restricting the dimensionality of z to be lower than that of ${x_1}$, the output of the encoder network that z represents is under complete representation of ${x_1}$. Secondly, a feed-forward network is used as a decoder to predict $\tilde{x_1}$ from z. Lastly, the loss function compares $\tilde{x_1}$ with its reference ${x_2}$. After training the model, bottleneck embedding z can be obtained from the encoder network of the trained model to measure neural entrainment distance. Unlike classical VAE, which reconstructs the same vector in compressed form, the bottleneck embedding z contains relevant entrainment information of ${x_1}$ and ${x_2}$.

We trained two DNN models with different auditory features for measuring auditory entrainment: LLD and TRILL. Both models have the same architecture. The architecture comprises two fully connected (FC) hidden layers in both the encoder and decoder networks. Both networks use batch normalization layers and Rectified Linear Unit (ReLU) activation layers between fully connected layers. The bottleneck embedding dimension is 30. For the LLD model, Hidden layers contain the following neuron units: [228 $\rightarrow$128 $\rightarrow$30 $\rightarrow$128 $\rightarrow$228]. We used the smooth L1 norm as the loss function and the Adam optimizer. For the TRILL model, the number of neuron units in the hidden layers is: [512 $\rightarrow$128 $\rightarrow$30 $\rightarrow$128 $\rightarrow$512]. We used the Kullback-Leibler (KL) divergence equation (\ref{equation:eq1}) as the loss function, and the Adam optimizer was applied. The number of epochs was set to 10. A 10-fold cross-validation technique was used to evaluate the models with a batch size of 128.
\begin{align}
KL({\tilde{x_1}},{x_2})={x_2}\cdot \log(\frac{{x_2}}{\tilde{x_1}})={x_2} \cdot (\log {x_2}-\log \tilde{x_1})
\label{equation:eq1}
\end{align}

\subsubsection{Semantic modelling}
\hspace{\parindent} We propose a different architecture for measuring semantic entrainment. Instead of learning representation from $\tilde{x_1}$ to ${x_2}$, we used auto-encoder architecture where loss from ${x_1}$ to $\tilde{x_1}$ is calculated as \textit{reconstruction loss}. Secondly, we also measure the loss on $\tilde{x_1}$ to ${x_2}$ as \textit{mapping loss} as shown in Figure \ref{fig:DNN_aud_sem} (b). Summing the two losses and backpropagation allows the model to learn semantic entrainment of the next turn based on the previous turn. These bottleneck features determine the Neutral Entrainment Distance (NED), a semantic entrainment measure.  

For semantic entrainment, we trained two DNN models: BERT and USE, for each dataset. The DNN models share the same auto-encoder architecture. Two fully connected (FC) hidden layers in both encoder and decoder networks. Between fully connected layers, both networks use batch normalization layers and Rectified Linear Units (ReLU). The bottleneck embedding size for the BERT model is 384, whereas the USE model is 30. For the BERT model, Hidden layers contain the following neuron units: [768 $\rightarrow$512 $\rightarrow$384 $\rightarrow$512 $\rightarrow$768]. Similarly, the USE model the size is: [512 $\rightarrow$128 $\rightarrow$30 $\rightarrow$128 $\rightarrow$512]. We used the smooth L1 norm as the loss function and the Adam optimizer in both models. A 10-fold cross-validation technique was used to evaluate the models with a batch size of 128.

\subsection{Neural entrainment distance (NED) measure}
After training the model, inputting the features into the encoder of the trained model will provide the bottleneck embeddings of the DNN model. The NED between these bottleneck embeddings is measured using different methods. 

\textit{Auditory model}: For the LLD model, we measure NED using the absolute difference between two bottleneck embeddings (say x1 and x2), as shown in equation (\ref{equation:eq2}). For the TRILL model, we measure NED using cosine distance between two bottleneck embeddings, as shown in equation (\ref{equation:eq3}). 

\textit{Semantic model}: For both BERT and USE models, we used cosine distance Eq.(\ref{equation:eq3}) as NED.

\begin{align}
NED_{lld} =\mid x_1-x_2 \mid
\label{equation:eq2}
\end{align}

\begin{align}
NED_{cosine\ distance}=\cos (x_1,x_2) = \frac{x_1 \cdot x_2}{|x_1||x_2|}
\label{equation:eq3}
\end{align}

\section{Results and Discussion} 
In order to validate NED as a valid proximity metric for entrainment, we ran three different classification experiments. The steps to measure the performance of each model are as follows: We measure two sets of distances for the entire session, namely consecutive NED and non-consecutive NED. Consecutive NED is the distance between two consecutive turns of speakers A and B. Non-consecutive NED is the distance between a speaker's turn and another random non-consecutive turn of another speaker. For DNN embeddings, if the consecutive NED is greater than the non-consecutive NED, we can infer that people are getting closer to each other and model is classifying cases correctly. Similarly, if consecutive NED distance is lower for LLD features, we can infer that people are entraining to each other, and these cases are categorized correctly. Later, the model accuracy is traditionally calculated as the proportion of correct cases.

\begin{table}[h!]
\centering
\resizebox{8cm}{!}{
\begin{tabular}{ccccccc}
\cline{2-7}
\multicolumn{1}{c|}{}         & \multicolumn{2}{c|}{Columbia games corpus}                                     & \multicolumn{2}{c|}{VAC corpus}                                                         & \multicolumn{2}{c|}{Fisher corpus}                                                      \\ \hline
\multicolumn{1}{|c|}{Feature} & \multicolumn{1}{c|}{IPU}          & \multicolumn{1}{c|}{Turn}                  & \multicolumn{1}{c|}{IPU}                   & \multicolumn{1}{c|}{Turn}                  & \multicolumn{1}{c|}{IPU}                   & \multicolumn{1}{c|}{Turn}                  \\ \hline
\multicolumn{1}{|l|}{LLD}     & \multicolumn{1}{c|}{74.69($\pm 3.81)$}  & \multicolumn{1}{c|}{\textbf{74.98 ($\pm 3.83)$}} & \multicolumn{1}{c|}{\textbf{77.94} ($\pm 3.84)$}          & \multicolumn{1}{c|}{77.26 ($\pm 3.97)$} & \multicolumn{1}{c|}{83.94 ($\pm 0.13)$}          & \multicolumn{1}{c|}{\textbf{84.14 ($\pm 0.03)$}} \\ \hline
\multicolumn{7}{c}{a) Classification accuracy for IPU vs Turn in three datasets}                                                                                                                                                                                                                   \\
\multicolumn{1}{l}{}          & \multicolumn{1}{l}{}              & \multicolumn{1}{l}{}                       & \multicolumn{1}{l}{}                       & \multicolumn{1}{l}{}                       & \multicolumn{1}{l}{}                       & \multicolumn{1}{l}{}                       \\ \cline{2-7} 
\multicolumn{1}{l|}{}         & \multicolumn{1}{c|}{A to B}       & \multicolumn{1}{c|}{B to A}                & \multicolumn{1}{c|}{Spkr to Alexa}         & \multicolumn{1}{c|}{Alexa to Spkr}         & \multicolumn{1}{c|}{A to B}                & \multicolumn{1}{c|}{B to A}                \\ \hline
\multicolumn{1}{|l|}{LLD}     & \multicolumn{1}{c|}{72.29 ($\pm 3.25)$} & \multicolumn{1}{c|}{74.29 ($\pm 3.23)$}          & \multicolumn{1}{c|}{58.25 ($\pm 4.98)$} & \multicolumn{1}{c|}{78.87 ($\pm 2.12)$}          & \multicolumn{1}{l|}{84.13 ($\pm 0.12)$}          & \multicolumn{1}{l|}{80.23 ($\pm 0.07)$}          \\ \hline
\multicolumn{7}{c}{b) Classification accuracy by splitting into groups}                                                                                                                                                                                                                            \\
\multicolumn{1}{l}{}          & \multicolumn{1}{l}{}              & \multicolumn{1}{l}{}                       & \multicolumn{1}{l}{}                       & \multicolumn{1}{l}{}                       & \multicolumn{1}{l}{}                       & \multicolumn{1}{l}{}                       \\ \cline{2-7} 
\multicolumn{1}{c|}{}         & \multicolumn{1}{c|}{One RT}       & \multicolumn{1}{c|}{Ten RT}                & \multicolumn{1}{c|}{One RT}                & \multicolumn{1}{c|}{Ten  RT}               & \multicolumn{1}{c|}{One RT}                & \multicolumn{1}{c|}{Ten RT}                \\ \hline
\multicolumn{1}{|c|}{LLD}     & \multicolumn{1}{c|}{\textbf{74.98} ($\pm 3.83)$} & \multicolumn{1}{c|}{71.23 ($\pm 3.45)$}          & \multicolumn{1}{c|}{\textbf{77.26} ($\pm 3.97)$}          & \multicolumn{1}{c|}{76.83 ($\pm 2.11)$}          & \multicolumn{1}{c|}{84.14 ($\pm 0.03)$}          & \multicolumn{1}{c|}{80.16 ($\pm 0.08)$}          \\ 
\multicolumn{1}{|c|}{TRILL}   & \multicolumn{1}{c|}{53.98 ($\pm$3.76)}             & \multicolumn{1}{c|}{56.17 ($\pm$5.82)}                      & \multicolumn{1}{c|}{31.17 ($\pm 3.67)$} & \multicolumn{1}{c|}{29.79 ($\pm 3.16)$} & \multicolumn{1}{c|}{94.14 ($\pm 0.01)$} & \multicolumn{1}{c|}{\textbf{93.75 ($\pm 0.02)$}} \\
\multicolumn{1}{|c|}{BERT}    & \multicolumn{1}{c|}{55.05 ($\pm 4.67)$} & \multicolumn{1}{c|}{\textbf{57.12 ($\pm 4.67)$}} & \multicolumn{1}{c|}{66.64 ($\pm 6.28)$}          & \multicolumn{1}{c|}{\textbf{67.55 ($\pm 5.21)$}} & \multicolumn{1}{c|}{60.31 ($\pm 0.04)$}          & \multicolumn{1}{c|}{61.64 ($\pm 0.04)$}          \\
\multicolumn{1}{|c|}{USE}     & \multicolumn{1}{c|}{47.03 ($\pm 3.35)$} & \multicolumn{1}{c|}{47.85 ($\pm 3.35)$}          & \multicolumn{1}{c|}{42.10 ($\pm 9.18)$}          & \multicolumn{1}{c|}{44.45 ($\pm 8.91)$}          & \multicolumn{1}{c|}{62.03 ($\pm 0.06)$}          & \multicolumn{1}{c|}{\textbf{63.05 ($\pm 0.06)$}} \\ \hline
\multicolumn{7}{c}{c) Classification accuracy for selecting different random turns (RT)} \\ 

\end{tabular}}

\caption{Summary of Classification accuracy for different auditory and semantic features on Columbia games corpus, Voice Assistant Conversation Corpus (VACC), and The Fisher Corpus English Part 1 (standard deviation shown in parentheses)}
    \label{tab:R1}
\end{table}

\subsection{\textbf{Experimnent 1: IPU vs Turn}} 
To analyze the effect of units of analysis on model performance, we trained six DNN entrainment models on three datasets using two units of analysis by extracting LLD features. Table \ref{tab:R1} (a) shows the accuracy of different units of analysis on three different datasets. In the Columbia games, VACC, and Fisher corpus, we found that models trained on the turn as a unit of analysis provide an accuracy of 74.98\%, 77.26\%, and 84.14\%, whereas models trained on IPU provide an accuracy of 74.69\%, 77.94\%, and 83.94\%. Hence, no significant difference in model performance trained on IPU and Turn in three datasets was found.

\subsection{\textbf{Experiment 2: Comparison in human-human and human-machine interaction}} To analyze if DNN auditory entrainment models can distinguish between HH and HM interaction we split each dataset into two groups: one has turns spoken by Speaker A first, followed by Speaker B, and the other has turns spoken by Speaker B, followed by Speaker A. We extracted LLD features from each group and trained the neural network models separately. Table \ref{tab:R1} (b) shows the accuracy of different groups on three different datasets. We found that the accuracy of the VACC corpus in the Speaker to Alexa group drops to 58.25\%  from 78.82\% in Alexa to Speaker. In the two HH corpora we did not find such a robust change between the groups. The drop in model performance of the VACC corpus can be explained as Alexa not entraining on auditory features with the speaker. Comparing consecutive and non-consecutive NED provides similar NED distance resulting in poor performance. We found DNN auditory models can distinguish between HH and HM interactions.     

\subsection{\textbf{Experiment 3: Comparison using different auditory and semantic embeddings}}
Table \ref{tab:R1} (c) shows the accuracy of different auditory and semantic entrainment models trained on different embeddings. To assess the robustness of the model, we measured non-consecutive distance in two ways. First, we used NED with one random turn (RT), and then we randomly selected ten non-consecutive turns of another speaker and calculated the mean NED. 

\subsubsection{\textbf{Experiment with Auditory features}} We compared the performance of LLDs and TRILL vectors on three datasets; rows 1-2 of Table \ref{tab:R1} (c). In the Columbia games and VACC corpora, LLD features outperformed TRILL vectors and provided higher accuracy of 74.98\% and 77.26\% when consecutive NED was compared to one random non-consecutive NED. Further, the performance dropped slightly when consecutive NED is compared to the mean of ten random non-consecutive NED on all three datasets. In contrast, TRILL vectors outperformed LLD features and provided higher accuracy of 94.14\% when consecutive NED was compared to one random non-consecutive NED in the Fisher corpus and 93.75\% when compared to the mean of ten random non-consecutive NED.  We also notice a drop in the performance of the LLD model by 4\% when consecutive NED was compared to the mean of 10 random NED, whereas there was a negligible drop of 0.39\% in the performance of TRILL vectors suggesting more robustness of the TRILL vector model.

The poor performance of TRILL vectors on two datasets (CGC and VACC) might be explained by the scarcity of data. TRILL vectors require a large amount of training data for learning representations in a meaningful way. Both datasets are smaller in size when compared to the Fisher corpus. The results reported in the current study on the Fisher Corpus are slightly worse than those reported in \cite{nasir2018a} where accuracy was reported at 98.87\%; we believe this is because we utilized a 10-fold cross-validation method instead of the hold-out method, which affects the performance of the model.
 
\subsubsection{\textbf{Experiment with Semantic features}} We train and evaluate the performance of variations of BERT (DistilBERT and  XLM-RoBERTa) and USE embeddings using DNN-based semantic models on three datasets. Rows 3-4 in Table \ref{tab:R1} (c) show that the Columbia Games corpus shows DistilBERT outperforms USE with an accuracy of 56.12 \% when compared to the mean of ten non-consecutive NED. Similarly, in the VACC corpus, we found that XLM-RoBERTa outperforms USE model with an accuracy of 67.55\% when consecutive NED is compared to the mean of ten random non-consecutive NED. In contrast, in the Fisher corpus, we found that the USE model provides slightly better accuracy of 63.05\% when compared to DistilBERT embeddings (62.03\%) when consecutive NED was compared to ten random non-consecutive turns. We also noticed an improvement in performance by 2\% in CGC and VACC corpora and 1\% in Fisher corpus when consecutive NED was compared to ten random turns.

Importantly, while training the models, we measured two losses: mapping loss and reconstruction loss. Without calculating the reconstruction loss, our novel contribution, accuracy dropped by 10\% and the model was not learning much. We can infer that semantic embeddings reduced by auto-encoders affect the model's performance. Secondly, we achieved the highest accuracy of 63.05\%. The language model for extracting semantic features can lead to this poor performance. Finally, the bottleneck embedding size affects the performance of the model. In auditory models, the bottleneck size was 30, whereas, in semantic models, the size is nearly half of the total embedding size. Reducing the size of bottleneck embeddings has a significant effect on the performance of the model.

We also conducted a qualitative error analysis to understand the poor performance of DNN semantic entrainment models on all three datasets. Table \ref{tab:R2} shows few instances on all three datasets where the model measured non-consecutive NED is smaller than consecutive NED. We observed that it is difficult for human readers to predict the next consecutive turn based on the choices shown in the table. We analyzed the Fisher corpus and the Columbia games corpus and found one probable cause of the errors is when backchannels and short utterances like okay, yeah, right are compared to measure NED. Both datasets comprise many affirmative cue words resulting in poor performance. In the VACC corpus, we observed the errors made by the model are caused when the model predicts consecutive speaker turn preceded by Alexa's turn. Since Alexa only responds to the questions asked by the speaker, the model fails to predict the next consecutive turn of the speaker.

\begin{table}[]
\centering
\resizebox{8cm}{!}{
\begin{tabular}{cccc}
\hline
\multicolumn{1}{|c|}{\textbf{Sr. No}} & \multicolumn{1}{c|}{{ Speaker A}}                                                                                           & \multicolumn{1}{c|}{{ Speaker B (non-consecutive)}}                                                                                                 & \multicolumn{1}{c|}{{ Speaker B (consecutive)}}                                                                                                                                                        \\ \hline
\multicolumn{1}{|c|}{1}               & \multicolumn{1}{c|}{{ and nail on the right okay}}                                                                          & \multicolumn{1}{c|}{{ \begin{tabular}[c]{@{}c@{}}and on the lower right a bottle of   \\ wine and a glass half full of wine\end{tabular}}} & \multicolumn{1}{c|}{{ \begin{tabular}[c]{@{}c@{}}right alien on the top yellow lion   \\ on the bottom left and money on \\ the bottom right\end{tabular}}}                                     \\ \hline
\multicolumn{1}{|c|}{2}               & \multicolumn{1}{c|}{{ on the top}}                                                                                          & \multicolumn{1}{c|}{{ I don't have anything like that}}                                                                                    & \multicolumn{1}{c|}{{ mmhm}}                                                                                                                                                                    \\ \hline
\multicolumn{1}{|c|}{3}               & \multicolumn{1}{c|}{{ eh I think it's worth taking}}                                                                        & \multicolumn{1}{c|}{{ oh okay yeah}}                                                                                                       & \multicolumn{1}{c|}{{ \begin{tabular}[c]{@{}c@{}}okay so I got outline of a airplane   \\ on the top um ball of yarn on the bottom \\ left and oreo   cookie on the bottom right\end{tabular}}} \\ \hline
                                      & \multicolumn{3}{c}{(a) Columbia games corpus}                                                                                                                                                                                                                                                                                                                                                                                                                                                                                          \\ \hline
\multicolumn{1}{|c|}{1}               & \multicolumn{1}{c|}{\begin{tabular}[c]{@{}c@{}}i don't know i just recovered \\ from a cold right now and uh\end{tabular}}                      & \multicolumn{1}{c|}{oh}                                                                                                                                        & \multicolumn{1}{c|}{okay}                                                                                                                                                                                           \\ \hline
\multicolumn{1}{|c|}{2}               & \multicolumn{1}{c|}{and so my strategy is do nothing}                                                                                           & \multicolumn{1}{c|}{right}                                                                                                                                     & \multicolumn{1}{c|}{okay}                                                                                                                                                                                           \\ \hline
\multicolumn{1}{|c|}{3}               & \multicolumn{1}{c|}{yeah that's true}                                                                                                           & \multicolumn{1}{c|}{oh wow}                                                                                                                                    & \multicolumn{1}{c|}{so how did you get into this study}                                                                                                                                                             \\ \hline
                                      & \multicolumn{3}{c}{(b) Fisher corpus English Part 1}                                                                                                                                                                                                                                                                                                                                                                                                                                                                                   \\ \hline
\multicolumn{1}{|c|}{1}               & \multicolumn{1}{c|}{\begin{tabular}[c]{@{}c@{}}Alexa book my appointment\\  on Tuesday the fifth\end{tabular}}                                  & \multicolumn{1}{c|}{\begin{tabular}[c]{@{}c@{}}Booking an appointment for \\ brainstorming IT at 2 p.m\end{tabular}}                                           & \multicolumn{1}{c|}{\begin{tabular}[c]{@{}c@{}}on Tuesday 5th December\\ no slots are available\end{tabular}}                                                                                                       \\ \hline
\multicolumn{1}{|c|}{2}               & \multicolumn{1}{c|}{\begin{tabular}[c]{@{}c@{}}On Monday fourth of \\ December there are four dates\end{tabular}}                               & \multicolumn{1}{c|}{Thanks Alexa}                                                                                                                              & \multicolumn{1}{c|}{\begin{tabular}[c]{@{}c@{}}Alexa what appointments do \\ I have on Tuesday the fifth\end{tabular}}                                                                                              \\ \hline
\multicolumn{1}{|c|}{3}               & \multicolumn{1}{c|}{\begin{tabular}[c]{@{}c@{}}On Wednesday December 6th \\ there are four appointments \\ at nine in the morning\end{tabular}} & \multicolumn{1}{c|}{\begin{tabular}[c]{@{}c@{}}Alexa, what slots are \\ available on wednesday\end{tabular}}                                                   & \multicolumn{1}{c|}{\begin{tabular}[c]{@{}c@{}}Alexa, book my appointments \\ on Wednesday\end{tabular}}                                                                                                            \\ \hline
                                      & \multicolumn{3}{c}{(c) VAC corpus}
\end{tabular}}
\caption{Error analysis for semantic entrainment on a) Columbia games corpus, b) The Fisher Corpus English Part 1,  and c) Voice Assistant Conversation Corpus (VACC)}
    \label{tab:R2}
\end{table}

\section{Conclusion} 
In this paper, we measured the performance of auditory and semantic features and trained DNN-based entrainment models using 10-fold cross-validation in HH and HMI corpora. We find that the unit of analysis (turn vs IPU) doesn't affect the performance of the DNN auditory entrainment model. Also, DNN entrainment models can distinguish between auditory entrainment in HH and HM interactions. Finally, we found TRILL vectors provide higher accuracy in the auditory entrainment model, and variation of BERT (XLM-RoBERTa) provides higher accuracy in DNN based semantic entrainment model.

\section{Acknowledgements}

\ifinterspeechfinal
     This project has received funding from the European Union’s Horizon 2020 \textit{research and innovation programme under the Marie Skłodowska-Curie grant agreement No} 859588 and in part from the Slovak Granting Agency grant VEGA2/0165/21 and Slovak Research and Development Agency grant APVV-21-0373.
\else
     
\fi

\bibliographystyle{IEEEtran}
\bibliography{mybib}

\end{document}